# Mevaker: Conclusion Extraction and Allocation Resources for the Hebrew Language

Vitaly Shalumov
vitaly.shalumov@gmail.com

Harel Haskey
harelnh@gmail.com

Yuval Solaz
yuval.solaz@gmail.com

March 10, 2024

## Abstract

In this paper, we introduce summarization **MevakerSumm** and conclusion extraction **MevakerConc** datasets for the Hebrew language based on the State Comptroller and Ombudsman of Israel reports, along with two auxiliary datasets. We accompany these datasets with models for conclusion extraction (**HeConE**, **HeConEspc**) and conclusion allocation (**HeCross**). All of the code, datasets, and model checkpoints used in this work are publicly available[1].

## 1 Introduction

While resources for pre-training language models for the Hebrew language continue to grow, mainly thanks to multi-lingual datasets scraped for multi-lingual model training, Hebrew resources for downstream tasks remain scarce. While tasks such as sentiment analysis, named entity recognition, and question answering received some attention (Amram et al., 2018), (Bareket and Tsarfaty, 2021), (Cohen et al., 2023), tasks such as summarization remain fairly untouched.

A close relative of the summarization task is conclusion extraction. While in summarization we are interested in extracting or generating spans that capture the context best, in conclusion extraction we are interested in extracting a higher logical knowledge level which merges the context with a prior knowledge of the author.

While most extractive summarization datasets can be best described as "take the first and last paragraph and you are good to go", for example, the CNN/Daily Mail dataset (Nallapati et al., 2016), a general purpose extractive summarization use-case is more similar to the WikiHow dataset (Koupaee and Wang, 2018), in which the extractive span is interwoven throughout the context.

Continuing the line established in (Shalumov and Haskey, 2023), we sought to extend the Hebrew NLP resources for both standard-length documents and long documents with emphasis on tasks with scarce resources. To that purpose, we turned to the State Comptroller and Ombudsman of Israel reports[2].

The State Comptroller and Ombudsman of Israel reports describe periodic State audits. State audit applies to a broad range of bodies from the public sector, among which are government ministries, State institutions, local authorities, statutory corporations, government companies, and additional agencies.

During the course of this work, these reports were processed to obtain two datasets: The first is an abstractive summarization dataset **MevakerSumm** which contains the context of the audit and its abstractive summary. The second is an extractive conclusions dataset **MevakerConc** which contains the context of the audit, the offsets of conclusions as marked by the auditors and the conclusions text contained within the offsets.

One of the goals of this paper is not only to provide additional datasets to the research community but also to provide additional models for scarcely researched tasks. Thus, we focused on two main tasks - Conclusion Extraction and Conclusion Allocation. To facilitate training the appropriate models, we synthesized two auxiliary datasets from Mevak-erConc: The **MevakerConcSen** dataset in which each sample contains a sentence, a label whether this sentence is a part of the conclusion, and a topic from which it was harvested. The second auxiliary dataset named **MevakerConcTree** is built for the conclusion allocation training process.

---

[1]The resources are available at the huggingface hub under the repository HeTree.

[2]https://www.mevaker.gov.il/sites/DigitalLibrary/Pages/Publications.aspx

Following the synthesis of the datasets, we trained several classifiers for conclusion extraction. This task can be achieved by numerous ways scaling in the complexity from single sentence classification through a sentence with a context around it classification up to seq2seq classification of sentence embeddings (Lukasik et al., 2020). We opted for training several models (**HeConE**, **HeConEspc**) with varying architecture complexity.

For the conclusion allocation task, we experimented with several approaches and finally chose to train a cross-encoder similarity model (**HeCross**) in which the unlabeled conclusions serve as query and hierarchical table of context concatenated with labeled conclusions serve as the paragraph.

This work introduces the following contributions:

- Synthesizing the abstractive summarization **MevakerSumm** and conclusion extraction **MevakerConc** datasets for the Hebrew language based on the State Comptroller and Ombudsman of Israel reports.

- Structuring the auxiliary conclusion allocation **MevakerConcTree** and sentence-level conclusion extraction **MevakerConcSen** datasets.

- Introducing the **HeConE** and **HeConEspc** models for conclusion extraction and the **HeCross** model for conclusion allocation.

- Publicly releasing the introduced datasets and models, all for the Hebrew language[3]

## 2 Datasets

The datasets for summarization and conclusion extraction are based on the State Comptroller and Ombudsman of Israel reports. We focused on the years 2005-2022 for which we downloaded the documents describing the audit and the hierarchical table of context of the audit. We focused on documents that could be automatically parsed for conclusions extraction, thus retaining 1109 documents. The document processing pipeline was based on automatic parsers such as PyPDF2[4], thus some parasitic attributes such as footnotes were retained. For illustration, a crop from one of the audit documents is given in Fig. 1.

---
[3]https://huggingface.co/HeTree

**הבטחת תקינות המבנים היבילים:** האחריות הישירה לתקינות המבנים המתקנים והתשתיות במוסדות החינוך ולעמידתם בדרישות החוק מוטלת על הרשות המקומית או על הבעלות של מוסד החינוך[36]. חוזר מנכ"ל משנת 2013 מפרט את הכללים וההנחיות להתקנת מבנים יבילים בבתי הספר ובגני הילדים, ובהם ההנחייה שהתקנת מבנים יבילים חייבת לעמוד בדרישות מפרט מכון התקנים.

בבקשת הרשות המקומית ממשרד החינוך למימון הצבת מבנה יביל יש מתחייבת להפעילו רק למטרות חינוך (כיתות לימוד או חדרי ספח), ואסור לה לשנות את ייעודו ללא הסכמת משרד החינוך מראש ובכתב.

עלה בביקורת כי אין למינהל הפיתוח נתונים עדכניים על המבנים היבילים שהוא תקצב ושהוקמו בפועל מבחינת סוג השימוש שנעשה בהם לאורך השנים, רמת האיכות שלהם ומקום הימצאם בפועל.

Figure 1: A crop from an audit document. The conclusion is bounded by the azure bounding box.

### 2.1 MevakerConc

Each of the 1109 audit documents contains conclusions made by the author of the report, marked by a specific bounding box. Thus, any text contained in the specific bounding box was labeled as a conclusion. Each of the 1109 data samples contains a context (document text), offsets of the conclusions which are woven throughout the context, the text contained within the offsets, and the topic which is the hierarchical heading of the document. Note that the majority of the documents also contain an abstractive summary, which was removed from the document. An example of a sample is given in Tab. 1

**Auxiliary Datasets** To facilitate the training process, we synthesized two auxiliary datasets from MevakerConc:

- MevakerConcSen: A sentence-level dataset that provides a label of conclusion/not conclusion (1/0 respectively) for each sentence together with indexes of the sentence and their document of origin.

---
[4]https://pypdf2.readthedocs.io

- MevakerConcTree: A dataset intended for the conclusion allocation task. The dataset represents several states of pre-allocated conclusions to a given hierarchical heading structure. Each sample contains a conclusion (Conc3 in Fig. 3) and a tree representation which is a concatenation of the hierarchical heading (for example concatenation of H1 and H11 in Fig. 3) of the documents with several other conclusions from the same document (Conc1 or Conc2 in Fig. 3). In the remainder of the paper, we will address the depth concatenation of all headlines and pre-allocated conclusions in the branch up until the deepest leaf as *branch*. The dataset was constructed as follows: given all conclusions in the document (N) and a bare tree structure (aggregated headlines of the table of context), we randomly selected one conclusion with the label being the bare tree. To construct the next sample, we add this conclusion to the bare tree and select a random conclusion from the remaining N-1 conclusions. This procedure continued until all conclusions for every document were allocated.

  Fig. 3

## 2.2 MevakerSumm

Recall that in MevakerConc construction we removed the abstractive summary from the document. By splitting the document into the aforementioned summary and the context (the same context as in MevakerConc), we created an abstractive summarization dataset for long documents. We removed from the original 1109 documents the ones that did not contain abstracts, thus retaining 1106 data samples.

## 3 Downstream Tasks

To further expand the resources available to the Hebrew language NLP researcher, we trained several models based on the MevakerConc dataset. HeConE and HeConEspc were trained for the task of conclusion extraction while HeCross was trained for the conclusion allocation task. We leave summarization model training based on the MevakerSumm dataset for future research.

### 3.1 Conclusion Extraction

This task can be addressed in numerous ways, as described in the Introduction section, thus we opted for training two extraction models, each with different solution architecture, and comparing the results.

The first model (HeConEspc) solves a simple binary sequence classification task in which each sample is a sentence with a given label (conclusion/not conclusion) aggregated with a context around it (window of $N$ sentences before and $N$ sentences after). While this sliding window approach is compute-intensive (a sample is constructed for each sentence), its architecture simplicity can facilitate the adoption of the model by researchers. However, a clear disadvantage of this model is the lack of dependency between the prediction of consecutive sentences that can improve sentence-level prediction. In addition, care must be taken when splitting to train and test as to avoid data leakage.

To improve the computational and dependency drawbacks of the HeConEspc model, but avoid the complication of introducing sentence embeddings as in (Lukasik et al., 2020), we reformulated the conclusion extraction task as token classification and trained the model (HeConE) as follows. Given a context window of $N$ sentences, a special token is inserted at the end of each sentence. This context window serves as an input while the classifier head classifies whether each token is a conclusion or not. By marking all but the special token irrelevant for loss calculations, we remain with loss terms from only the special tokens prediction. Fig. 2 illustrates the model architecture.

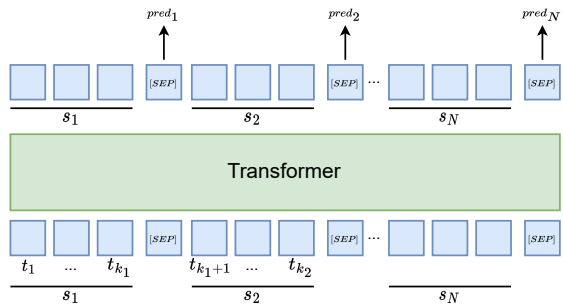

Figure 2: HeConE model architecture.

Table 1: A sample from the **MevakerConc** dataset.

| Context | Conclusions | Offsets | Topic |
|---|---|---|---|
| החברה הממשלתית לתיירות (להלן - החברה) הוקמה בשנת 1955 ועסקה במכלול פעילויות התיירות עד הקמתו של משרד התיירות בשנת 1964 . עם הקמת משרד התיירות, החלה החברה לשמש זרוע ביצועי שלו לפיתוח תשתיות תיירות. לצורך ביצוע פעילותה מסתייעת החברה בשירותים של קבלנים ובעלי מקצוע מומחים, לרבות מומחים בתחומי התכנון, ההנדסה, הפיקוח וניהול פרויקטים (להלן - יועצים חיצוניים). החברה מבצעת פרויקטים בהיקף של כ-80 מיליון ש"ח בשנה ותקציב המינהלה שלה הוא כ-7.5 מיליון ש"ח בשנה. תקציב הפרויקטים מתחלק בין פרויקטים בביצוע החברה ופרויקטים בביצוע שותפים אחרים .... | מבדיקת משרד מבקר המדינה עולה כי החברה העסיקה את היועץ ... | [4111, 4393] | החברה הממשלתית לתיירות/ העסקת יועצים - החברה הממשלתית לתיירות בע"מ |
| | לדעת משרד מבקר המדינה היה על החברה לקבל החלטה בפרק זמן... | [5331, 5691] | |
| | מבדיקת משרד מבקר המדינה עולה כי התעריף המתאים ליועץ ... | [8072, 8380] | |
| | משרד מבקר המדינה מעיר לחברה כי לא היה מקום לשלם עבור ... | [9383, 9731] | |
| | מבדיקת משרד מבקר המדינה עולה כי היה על החברה להפחית ... | [10309, 10445] | |
| | משרד מבקר המדינה מעיר לחברה כי בהתאם להוראת החשב הכללי ... | [10749, 11339] | |
| | מבדיקת משרד מבקר המדינה עולה כי תעריף החשב הכללי ... | [11767, 12150] | |
| | לדעת מ שרד מבקר המדינה, על החברה לבחון את נחיצות המשך ... | [14251, 14459] | |
| | מבדיקת משרד מבקר המדינה עולה כי החברה העסיקה חמישה ... | [16521, 16821] | |
| | משרד מבקר המדינה מעיר לחברה כי העסקת היועצים מיד לאחר ... | [17297, 17596] | |
| | משרד מבקר המדינה מעיר לחברה כי הצורך שלה בהמשך העסקתו ... | [18689, 18823] | |
| | מבדיקת משרד מבקר המדינה עולה כי ההתקשרות עם שבעת ... | [19260, 19376] | |
| | דוח זה מצביע על ליקויים בתהליך בחירת יועצים חיצוניים ... | [19383, 20073] | |

## 3.2 Conclusion Allocation

We formulate the conclusion allocation task as assigning a conclusion (typically several sentences long) to the deepest headline of the underlining table of context of the reports. To facilitate a more general case usage, we assume that several conclusions may be already pre-assigned to a specific headline, thus enriching the heading with additional context. The conclusion allocation task is exemplified in the Fig. 3.

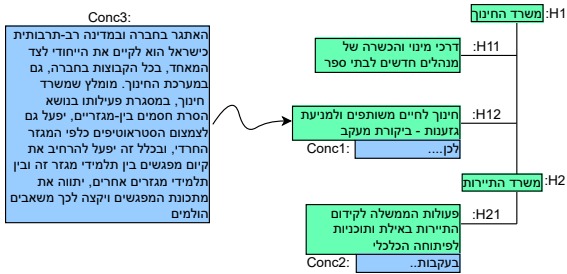

Figure 3: Conclusion allocation task.

We experimented with supervised and unsupervised approaches approaches as follows.

**Question Answering** This approach treats the problem as question answering where the question is the branch and the leaf of the context tree (for example H1+H11 from Fig. 3) and the context is the conclusion. The decision of whether the question can be answered from the context was determined by a predefined threshold. We used the HeRo model fine-tuned on the ParaShoot dataset (Keren and Levy, 2021) as our question-answering model. While this approach demonstrated robustness to the diversity of the topics, it missed several obvious conclusion allocations. Thus, we opted to abandon this formulation of the problem.

**Similarity** The most straightforward approach to the task of conclusion allocation to any tree is using the similarity between a context and a branch. The two dominating types of models in the similarity domain are the bi-encoder and cross-encoder. While bi-encoder models excel at fast retrieving by the utilization of first representing each text as an embedding vector and then calculating similarity using a simple measure such as dot product, a more accurate similarity evaluation can be achieved by the cross-encoder model that relies on the text itself and not its compact representation. While a dedicated bi-encoder or cross-encoder for the Hebrew language did not exist at the time of writing this paper, several multi-lingual bi-encoder models that include Hebrew are available[5]. Thus, we chose to train a dedicated cross-encoder model for the Hebrew language (HeCross) and utilize it for conclusion allocation.

---
[5]https://huggingface.co/sentence-transformers/paraphrase-xlm-r-multilingual-v1 by (Reimers and Gurevych, 2019) and https://huggingface.co/intfloat/multilingual-e5-large by (Wang et al., 2022).

# 4 Core Results

## 4.1 Conclusion Extraction

We fine-tuned the HeRo Hebrew language model(Shalumov and Haskey, 2023) on the MevakerConcSen dataset using both a special token (HeConE) and a sliding window (HeConEspc) approaches for 20 epochs with the default Huggingface (Wolf et al., 2020) parameters. To avoid leakage between the train and test splits in the HeConEspc training due to overlapping windows in train and test, we opted to disable shuffling for said training. We used a window length of $w_{HeConE} = 5$ and $w_{HeConEspc} = 5$ for the models. In the HeConE, $w$ represents how many sentences are aggregated to a single input (i.e. the $N$ value in Fig. 3. For the HeConEspc, $w$ is the number of sentences used for context ($w$ before and $w$ after). Note that the sliding window approach takes roughly $w_{HeConE}$ more computational time than the special token approach due to the overlapping windows.

**Balancing the Dataset.** The MevakerConcSen dataset is highly unbalanced due to the nature of the parsed reports. We chose to balance the train splits of the dataset for each model training as follows (the test split was untouched) :

- HeConE: We artificially duplicated the windowed samples that have more conclusions sentences than sentences that are not conclusions until a similar conclusion/not conclusion total count was reached.

- HeConEspc: We artificially duplicated the windowed samples that were labeled as conclusions together with their appropriate context until a similar conclusion/not conclusion total count was reached.

The results for the conclusion extraction task on the internal split of the datasets are given in Tab. 2.

## 4.2 Conclusion Allocation

To increase the diversity of the similarity dataset, we have added samples from the HeQ question answering dataset (Cohen et al., 2023) to the MevakerConcTree dataset. The similarity samples for HeQ were created by labeling as similar the question and the paragraph in which the answer lies. Dissimilar samples were obtained by randomly taking other paragraphs from the dataset.

Table 2: F1 results on the MevakerConcSen dataset.

| Model | MevakerConcSen |
|---|---|
| HeConE | 84.10 |
| HeConEspc | 90.83 |

We fine-tuned the HeRo model using the sentence-transformers library[6] for a single epoch of 200k samples (100k from HeQ and 100k from MevakerConcTree).

To evaluate the model we turned to baseline models in the English language. The evaluation of the model was performed using the parallel sentences dataset (PS) which provides an English sentence with its translation on a target language (Reimers and Gurevych, 2020). We took the "dev" split of the English-Hebrew sentences (1000 sentences), and created all permutations of sentence pairs, thus acquiring 999000 tuples for both English and Hebrew. Next, we ranked the pairs by the similarity score obtained by the English cross-encoder model[7] which was treated as ground truth. To the aforementioned ground truth, we compared several bi-encoder models introduced in (Wang et al., 2022) and our cross-encoder model. When applicable, the evaluated model was used for both English and Hebrew similarity ranking.

The evaluation was performed on two metrics - Kendall Rank Correlation (KRC) (Kendall, 1938) and Mean Absolute Error (MAE) normalized by number of samples. These metrics were preferred over retrieval metrics that discount lower rankings which is not applicable to our evaluation use case. We chose to compare relative rankings and not absolute similarity scores to establish common ground between models. The results are given in Tab. 3 for a random sample of 10k tuples.

Tab. 3 demonstrates that when comparing bi-encoders, the best results are obtained by a dedicated English bi-encoder (e5-large-v2). The multi-lingual bi-encoder achieves similar results on both the English and the Hebrew data samples. Interestingly, the results of the multilingual bi-encoder are slightly higher in Hebrew than in English data samples. The introduced Hebrew cross-encoder surpasses both English and multi-lingual bi-encoders making it a viable option for similarity evaluation in the Hebrew language, especially if computational time restrictions permit it.

---

[6]https://www.sbert.net/
[7]https://huggingface.co/cross-encoder/ms-marco-MiniLM-L-12-v2

Table 3: Similarity ranking results compared to ms-marco-MiniLM-L-12-v2 cross-encoder. *cross* and *bi* represent cross-encoder and bi-encoder architectures, while *en* and *he* stand for prediction on the English and the Hebrew data samples respectively. KRC is between -1 (dissimilar) and 1 (similar) - higher is better. For MAE lower is better.

| Model | KRC | MAE |
|---|---|---|
| e5-large-v2 (bi, en) | 0.0094 | 0.3331 |
| multilingual-e5-large (bi, en) | 0.0006 | 0.3341 |
| multilingual-e5-large (bi, he) | 0.0338 | 0.3257 |
| HeCross (cross, he) | **0.0647** | **0.3142** |

## 5 Conclusions

In this paper, we continued to expand the available resources for the Hebrew NLP community. Our We focused on resources for two main tasks - Conclusion Extraction and Conclusion Allocation. We initialized our research with the processing of the State Comptroller and Ombudsman of Israel reports, from which we constructed several datasets: Mevaker-Summ for abstractive summarization, MevakerConcTree for conclusion allocation, Mevaker-Conc for conclusion extraction, and Mevaker-ConcSen for sentence-level conclusion extraction.

We then proceeded to train several models for the conclusion extraction task (HeConE, HeConEspc) with different classification architectures. Finally, we trained the first monolingual cross-encoder similarity model for the Hebrew language named HeCross.

We publicly release all datasets and models under the Huggingface repository HeTree[8].

---

[8] https://huggingface.co/HeTree

## 6 Acknowledgements

We are grateful to Tal Geva and Amir David Nissan Cohen for technical discussions during the project.